# Placement Optimization with Deep Reinforcement Learning

Anna Goldie and Azalia Mirhoseini
agoldie,azalia@google.com
Google Brain

## ABSTRACT

Placement Optimization is an important problem in systems and chip design, which consists of mapping the nodes of a graph onto a limited set of resources to optimize for an objective, subject to constraints. In this paper, we start by motivating reinforcement learning as a solution to the placement problem. We then give an overview of what deep reinforcement learning is. We next formulate the placement problem as a reinforcement learning problem, and show how this problem can be solved with policy gradient optimization. Finally, we describe lessons we have learned from training deep reinforcement learning policies across a variety of placement optimization problems.

## 1 INTRODUCTION

An important problem in systems and chip design is Placement Optimization, which refers to the problem of mapping the nodes of a graph onto a limited set of resources to optimize for an objective, subject to constraints. Common examples of this class of problem include placement of TensorFlow graphs onto hardware devices to minimize training or inference time, or placement of an ASIC or FPGA netlist onto a grid to optimize for power, performance, and area.

Placement is a very challenging problem as several factors, including the size and topology of the input graph, number and properties of available resources, and the requirements and constraints of feasible placements all contribute to its complexity. There are many approaches to the placement problem. A range of algorithms including analytical approaches [3, 12, 14, 15], genetic and hill-climbing methods [4, 6, 13], Integer Linear Programming (ILP) [2, 27], and problem-specific heuristics have been proposed.

More recently, a new type of approach to the placement problem based on deep Reinforcement Learning (RL) [16, 17, 28] has emerged. RL-based methods bring new challenges, such as interpretability, brittleness of training to convergence, and unsafe exploration. However, they also offer new opportunities, such as the ability to leverage distributed computing, ease of problem formulation, end-to-end optimization, and domain adaptation, meaning that these methods can potentially transfer what they learn from previous problems to new unseen instances.

In this paper, we start by motivating reinforcement learning as a solution to the placement problem. We then give an overview of what deep reinforcement learning is. We then formulate the placement problem as an RL problem, and show how this problem can be solved with policy gradient optimization. Finally, we describe lessons we have learned from training deep RL policies across a variety of placement optimization problems.

## 2 DEEP REINFORCEMENT LEARNING

Most successful applications of machine learning are examples of supervised learning, where a model is trained to approximate a particular function, given many input-output examples (e.g. given many images labeled as cat or dog, learn to predict whether a given image is that of a cat or a dog). Today's state-of-the-art supervised models are typically deep learning models, meaning that the function approximation is achieved by updating the weights of a multi-layered (deep) neural network via gradient descent against a differentiable loss function.

Reinforcement learning, on the other hand, is a separate branch of machine learning in which a model, or policy in RL parlance, learns to take actions in an environment (either the real world or a simulation) to maximize a given reward function. One well-known example of reinforcement learning is AlphaGo [23], in which a policy learned to take actions (moves in the game of Go) to maximize its reward function (number of winning games). Deep reinforcement learning is simply reinforcement learning in which the policy is a deep neural network.

RL problems can be reformulated as Markov Decision Processes (MDPs). MDPs rely on the Markov assumption, meaning that the next state $s_{t+1}$ depends only on the current state $s_t$, and is conditionally independent of the past.

$$P(s_{t+1}|s_0...s_t) = P(s_{t+1}|s_t)$$

Like MDPs, RL problems are defined by five key components:

- states: the set of possible states of the world (e.g. the set of valid board positions in Go)
- actions: the set of actions that can be taken by the agent (e.g. all valid moves in a game of Go)
- state transition probabilities: the probability of transitioning between any two given states.

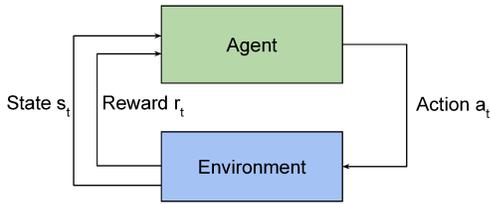

Figure 1: In Reinforcement Learning, an agent explores the environment and finds actions to maximize cumulative rewards.

- reward: the objective to be maximized, subject to future discounting as defined below (e.g. 1 if you win the game of Go, 0 otherwise)
- discount for future rewards: how much to discount the value of future reward, due to its relative uncertainty (e.g. a discount factor of .95 would mean that 95 dollars today is equivalent to 100 tomorrow).

At each time step t, the agent begins in state ($s_t$), takes an action ($a_t$), arrives at a new state ($s_{t+1}$), and receives a reward ($r_t$) from the environment, as shown in Figure 1.

Through repeated episodes (sequences of states, actions, and rewards), the agent learns to take actions that will maximize cumulative reward.

Reinforcement learning approaches can be divided into two broad categories: model-free and model-based. In model-free reinforcement learning, we train a policy to take actions that maximize reward from a black-box environment. In model-based reinforcement learning, we train a policy to take actions that maximize reward, while also training an explicit model of the world, which learns to predict the reward and state transitions of the environment. Most existing work on reinforcement learning for systems problems has taken a model-free approach, as it is generally easier to train to convergence. However, model-based reinforcement learning has been shown to be more sample efficient in other domains [11], so it may be a viable direction to take in future work, especially in situations where the reward function is very expensive to evaluate.

Since the agent's goal is to maximize cumulative reward, one approach is to learn a value function that can predict the reward given a state, $v(s)$, and then take the action which will bring the agent into a state that yields the highest reward. However, a more common approach in recent years is to use policy gradient methods, which seek to directly learn the policy $\pi(a|s)$ that predicts the optimal action given the current state. Popular policy gradient methods include REINFORCE [25], A3C [18], TRPO [21], and PPO [22].

RL is helpful in cases where we do not have sufficient labeled data (input-output examples) to take a supervised learning approach or when the objective function is not differentiable. It is also well-suited to massive search problems, where exhaustive or heuristic-based methods cannot scale, such as AlphaGo [23], AlphaStar [24], and OpenAI Five DOTA [19].

Reinforcement learning policies are famously difficult to train, as they tend to be brittle with respect to their hyperparameters, hard to interpret and debug, and prone to catastrophic failures and unsafe exploration. This area of machine learning is not yet as well understood, with few educational resources available, as compared to more established areas like deep learning. This is part of our motivation for writing this paper, to demystify this area and empower more systems experts to move into this promising, but challenging area.

## 3 DEEP REINFORCEMENT LEARNING FOR PLACEMENT OPTIMIZATION

In this section, we first start by formulating the placement problem. We will then show how RL can be applied to solve this problem.

### 3.1 Placement Problem Formulation

Let us assume the input graph $g$ has nodes $v_1, v_2, ..., v_N$. We want to place these nodes onto placement locations $l_1, l_2, ..., l_M$. In this set up, we use RL to find a mapping

$$(v_1, v_2, ..., v_N) \rightarrow (l_1, l_2, ..., l_M)$$

that maximizes a reward function $R$ subject to constraints. Here, there is a unique placement location for each node $v_i$, but each location $l_j$ can be assigned to multiple nodes. The constraints vary by problem, but a common constraint is limited capacity for each placement location, meaning that there is a limit on how many nodes can be assigned to each location. In the next section, we will discuss some case-specific constraints and ways to incorporate them.

To formulate placement as a policy optimization problem, [17] proposed relaxing the objective function. Instead of finding the absolute best placement, one can train a policy that generates a probability distribution of nodes to placement locations such that it maximizes *the expected reward* generated by those placements.

Let us denote the policy $\pi$ parameterized by $\theta$ as $\pi_\theta$. $\theta$ represents the weights of a deep network architecture. Here, we describe the objective, which is to train parameters $\theta$ such that the network predicts placement decisions for the nodes of the input graph $g$, and as a result, the placement reward $R_{l,g}$ is maximized. We can write the cost function we are optimizing for as follows:

$$J(\theta, g) = E_{g, l \sim \pi_\theta}[R_{l,g}] = \sum_{l \sim \pi_\theta} \pi_\theta(l|g) R_{l,g} \quad (1)$$

We can then train this policy (optimize parameters $\theta$) using a policy gradient based method, which we will discuss in this section.

Here, we provide an overview of the state and action space, reward function and the neural network architecture of the policy by delving into example placement problems, namely TensorFlow device placement, ASIC netlist placement, and FPGA netlist placement. As shown in Figure 2, all of these placement problems require mapping the nodes of a graph onto placement locations such that their corresponding reward metrics are optimized. Target placement locations for TensorFlow graphs, ASIC netlists, and FPGA netlists are the computing devices (e.g., TPU or GPUs), grid cells of the chip canvas, and FPGA Configurable Logic Blocks (CLBs), respectively.

For each of these problems, the neural network policy receives a state as input, and outputs an action for that state. In general,

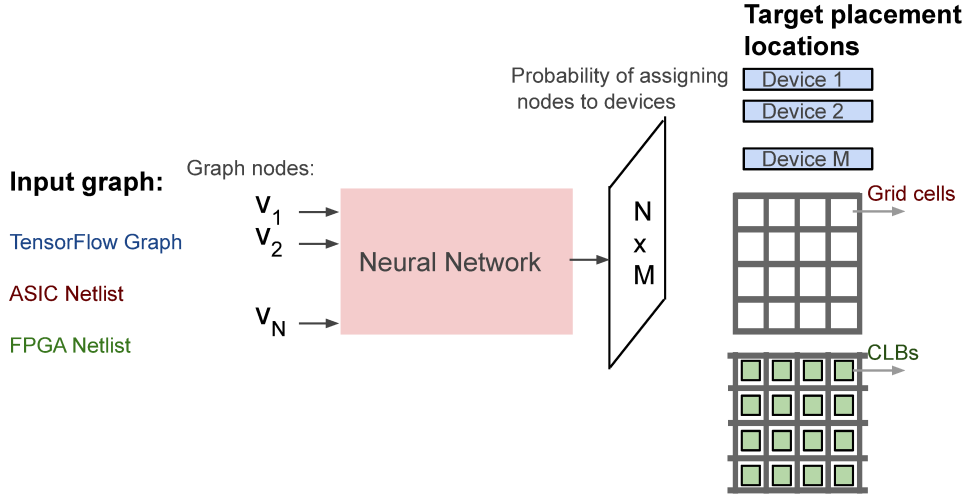

Figure 2: Placement optimization overview. Target placement locations for TensorFlow graphs, ASIC netlists, and FPGA netlists are the computing devices (e.g., TPU or GPUs), grid cells of the chip canvas, and FPGA Configurable Logic Blocks (CLBs), respectively.

the state should represent the information that the policy needs for prediction, such as node ID, type, and adjacency matrix. The network then outputs a probability distribution representing the probability of assigning an input node onto each placement location. The action is selected by sampling or taking the argmax of the output probability distribution.

The reward function varies for different problems. For example, for TensorFlow graph placement, we use negative runtime of a training step of the placed deep network model. For ASIC and FPGA netlists, the reward is more complex and should include various metrics related to power and timing (e.g. total wirelength, routability congestion, and cell density).

## 3.2 Graph Convolutional Neural Networks

As discussed, many placement problems take input in the form of graph. The way in which these input graphs are represented has great impact on the ability of machine learning models to generate high-quality placements. More meaningful representations also help models to learn patterns that generalize to new unseen graphs, as opposed to merely memorizing the graphs that they encounter. We therefore describe some of the recent advances in neural graph representations, such as Graph Convolutional Networks.

Graph neural networks can be divided into four high-level categories [26]: recurrent graph neural networks (RecGNNs) [7, 8, 20], convolutional graph neural networks (ConvGNNs) [1, 5, 10], graph autoencoders (GAEs), and spatial-temporal graph neural networks (STGNNs). RecGNNs preceded ConvGNNs and pioneered the idea of message passing, or representing a node as an iterative aggregate of its neighbors. Each iteration of message passing results in one additional hop (e.g. running the algorithm for k iterations would result in each node being influenced by all neighbors within a k-hop radius). As such, these methods better encode the overall topology of the graph, enabling domain adaptation, as described in the following section. Most graph neural network methods used in systems today are ConvGNNs, which generalize the concept of convolution. After all, images are merely a special case of graphs, in which pixels are nodes connected to the pixels (nodes) immediately surrounding them. ConvGNNs use deep convolutional networks to capture even distant relationships within the graph.

## 3.3 Domain Adaptation

Domain adaptation in placement is the problem of training policies that can learn across multiple graphs and transfer the acquired knowledge to generate more optimized placements for new unseen graphs. In the context of the three examples we discussed in this work, domain adaptation means we train a policy across a set of TensorFlow graphs, ASIC or FPGA netlists and apply the trained policy to an unseen TensorFlow graph, ASIC or FPGA netlist. [28] proposed the problem formulation for a domain adaptive policy as follows:

$$J(\theta, G) = \frac{1}{K} \sum_{g \sim G} E_{g,l \sim \pi_\theta}[R_{l,g}] \quad (2)$$

In this case, $G$ is a set of $K$ training graphs. The training procedure can be similar to that of traditional machine learning (e.g., using a holdout validation set or cross validation).

## 3.4 Solving the Placement Objective with Policy Gradient Optimization

Now that we have defined an objective function, we will explain how to use *policy gradient optimization* to learn the parameters $\theta$ of the policy. We can write the derivative of the objective function in Equation 1 as follows:

$$\Delta_\theta J(\theta, g) = \sum_{l \sim \pi_\theta} \Delta_\theta \pi_\theta(l|g) R_{l,g} \quad (3)$$

$$= \sum_{l \sim \pi_\theta} \pi_\theta \frac{\Delta_\theta \pi_\theta(l|g)}{\pi_\theta} R_{l,g} \quad (4)$$

$$= \sum_{l \sim \pi_\theta} \pi_\theta \Delta_\theta log(\pi_\theta(l|g)) R_{l,g} \quad (5)$$

$$= E[\Delta_\theta log(\pi_\theta(l|g)) R_{l,g}] \quad (6)$$

The equation above is the basis of various policy gradient optimization methods, such as REINFORCE [25], PPO [22], and SAC [9].

## 4 INGREDIENTS FOR RL SUCCESS

In this section, we will share some of the lessons that we have learned in training deep reinforcement learning policies to solve placement problems in computer systems and chip design.

**Reward Function:** Designing the right reward function is one of the most critical decisions. Some properties of effective reward functions are as follows:

1) Reward functions should be fast to evaluate; RL training often requires 10-100s of thousands of iterations of reward evaluation before reaching convergence. While the exact timing that makes a tractable reward function depends on the complexity of the problem, a sub-second reward function would be effective in nearly any scenario.

2) Reward functions should be strongly correlated with the true objective. In many real-world scenarios, we need to use simulators or proxy reward functions to approximate the true objective, which may be prohibitively expensive to calculate. If the proxy reward is not well-correlated with the true objective, we are solving the wrong problem and the learned placement is unlikely to be useful. While designing a good simulator or approximate function is a challenging task in its own right, it is helpful to build a reward function by combining various approximate metrics that each independently correlate with the true reward. For example, for TensorFlow placement, the proxy reward could be a composite function of total memory per device, number of inter-device (and therefore expensive) edges induced by the placement, imbalance of computation placed on each device. By incorporating a weighted average of multiple proxy rewards, the total variance of the reward error is reduced and over-fitting to a particular proxy metric is avoided.

3) Another important factor is correctly engineering the reward function. This could be as simple as normalizing the reward or applying more complex functions to change the shape of the reward. For example, for the device placement problem, measuring the runtime of one step of the TensorFlow graph was the true reward function. Due to the runtime varying widely across different placements, using the runtime directly would interfere with learning and gradient updates. We chose to instead use the square root of the runtime, which effectively dampened the range of values.

**Action Space:** Another key ingredient is designing the appropriate action space. For example, the problem could be formulated as placing the nodes of the netlist one at a time onto the chip netlist, or as placing all of the nodes and then deciding which perturbation (e.g. swap, shift, rotate, etc.) to apply to each of the nodes in a fixed sequence. In device placement, we chose to place all of the TensorFlow nodes onto hardware devices before evaluating the reward for that placement, because otherwise measuring the reward of a partial placement would be very difficult, if not impossible. For ASIC placement, on the other hand, one can define partial reward functions, because it is possible to measure changes in metrics, such as wirelength and congestion, as nodes are being placed.

**Managing Constraints:** The constraints for feasible placements vary across placement problems. For example, a common constraint is the capacity of placement locations, which limits the number of nodes that can be placed onto that location. For example in device placement, the memory footprint of the nodes placed onto a single device should not exceed the memory limit of that device. Another constraint is that certain nodes cannot co-exist on the same location. For example, in ASIC placement, two macro blocks cannot overlap on the chip canvas.

There are many approaches to enforcing these constraints to avoid or reduce the number of infeasible placements generated by the policy. Perhaps the most straightforward way to handle the constraints is to penalize the policy with a large negative reward whenever it generates infeasible placements. A challenge with this solution is that the policy does not gain any information about how far this placement was from a feasible placement. In the most extreme case, if all of the initial placements generated by the policy are infeasible, there will be no positive signal to teach the policy how to explore the environment and training will fail. Thus, creating a reward function that penalizes the infeasible placements relative to how far they are from viable placements becomes critical.

Another approach is to force the policy to only generate feasible placements. This can be accomplished via a function that masks out the infeasible placements. For example, a mask can be updated given the partial placement of the graph nodes. Each time a new node is placed, the density of all the locations is updated (based on the locations of the nodes that are already placed). The action space then becomes limited to those locations that have enough free capacity to accept the new node. This approach has its own challenges as calculating the mask, similar to calculating the reward, must be done efficiently.

**Representations:** Finally, the way in which state is represented has significant impact on the performance of the policy and its ability to generalize to unseen instances of the placement problem. For example, in the earlier TensorFlow device placement papers [16, 17], we represented the computational graph as a list of adjacencies, indices of the node operations, and sizes of the operations. This approach was effective when the policy was trained from scratch for each new TensorFlow graph, but was unable to generalize or transfer what it learned to new graphs. On the other hand, [28] used graph convolutional neural networks to learn better representations of the computational graph structure, and were able to transfer knowledge across graphs.

## 5 CONCLUSION

In this paper we discuss placement optimization with deep reinforcement learning. Deep RL is a promising approach for solving combinatorial problems, and enables domain adaptation and direct optimization of non-differentiable objective functions. Training RL policies is a very challenging task, in part due to the brittleness of gradient updates and the costliness of evaluating rewards. In this work, we provide an overview of deep RL, formulate the placement problem as a RL problem, and discuss strategies for training successful RL agents.

We predict a trend towards more effective RL-based domain adaptation techniques, in which graph neural networks will play a key role in enabling both higher sample efficiency and more optimal placements. We also foresee a future in which easy-to-use RL-based placement tools will enable non-ML experts to harness and improve upon these powerful techniques.

## ACKNOWLEDGMENTS

We would like to thank our amazing collaborators on deep reinforcement learning for placement research, including Ebrahim Songhori, Joe Jiang, Shen Wang, Hieu Pham, Yanqi Zhou, Will Hang, Azade Nazi, Sudip Roy, Amir Yazdanbakhsh, Benoit Steiner, Rasmus Larsen, Yuefeng Zhou, Naveen Kumar, Mohammad Norouzi, Samy Bengio, Hieu Pham, James Laudon, Quoc Le, and Jeff Dean. We would also like to thank Gabriel Warshauer-Baker for reviewing the paper and improving its clarity.

## REFERENCES

[1] Joan Bruna, Wojciech Zaremba, Arthur Szlam, and Yann LeCun. 2013. Spectral Networks and Locally Connected Networks on Graphs. arXiv:cs.LG/1312.6203
[2] A. Chakraborty, A. Kumar, and D. Z. Pan. 2009. RegPlace: A high quality open-source placement framework for structured ASICs. In *2009 46th ACM/IEEE Design Automation Conference*. 442–447.
[3] C. Cheng, A. B. Kahng, I. Kang, and L. Wang. 2019. RePlAce: Advancing Solution Quality and Routability Validation in Global Placement. *IEEE Transactions on Computer-Aided Design of Integrated Circuits and Systems* 38, 9 (2019), 1717–1730.
[4] J. P. Cohoon and W. D. Paris. 1987. Genetic Placement. *IEEE Transactions on Computer-Aided Design of Integrated Circuits and Systems* 6, 6 (November 1987), 956–964. https://doi.org/10.1109/TCAD.1987.1270337
[5] Michaël Defferrard, Xavier Bresson, and Pierre Vandergheynst. 2016. Convolutional Neural Networks on Graphs with Fast Localized Spectral Filtering. arXiv:cs.LG/1606.09375
[6] H. Esbensen. 1992. A genetic algorithm for macro cell placement. In *Proceedings EURO-DAC '92: European Design Automation Conference*. 52–57. https://doi.org/10.1109/EURDAC.1992.246265
[7] C. Gallicchio and A. Micheli. 2010. Graph Echo State Networks. In *The 2010 International Joint Conference on Neural Networks (IJCNN)*. 1–8. https://doi.org/10.1109/IJCNN.2010.5596796
[8] M. Gori, G. Monfardini, and F. Scarselli. 2005. A new model for learning in graph domains. In *Proceedings. 2005 IEEE International Joint Conference on Neural Networks, 2005.*, Vol. 2. 729–734 vol. 2. https://doi.org/10.1109/IJCNN.2005.1555942
[9] Tuomas Haarnoja, Aurick Zhou, Pieter Abbeel, and Sergey Levine. 2018. Soft Actor-Critic: Off-Policy Maximum Entropy Deep Reinforcement Learning with a Stochastic Actor. arXiv:cs.LG/1801.01290
[10] Mikael Henaff, Joan Bruna, and Yann LeCun. 2015. Deep Convolutional Networks on Graph-Structured Data. arXiv:cs.LG/1506.05163
[11] Michael Janner, Justin Fu, Marvin Zhang, and Sergey Levine. 2019. When to Trust Your Model: Model-Based Policy Optimization. arXiv:cs.LG/1906.08253
[12] Myung-Chul Kim, Jin Hu, Dong-Jin Lee, and Igor L. Markov. 2011. A SimPLR Method for Routability-Driven Placement. In *Proceedings of the International Conference on Computer-Aided Design* (San Jose, California) *(ICCAD '11)*. IEEE Press, 67–73.
[13] S. Kirkpatrick, C. D. Gelatt, and M. P. Vecchi. 1983. Optimization by Simulated Annealing. *Science* 220, 4598 (1983), 671–680. https://doi.org/10.1126/science.220.4598.671 arXiv:https://science.sciencemag.org/content/220/4598/671.full.pdf
[14] Yibo Lin, Shounak Dhar, Wuxi Li, Haoxing Ren, Brucek Khailany, and David Z. Pan. 2019. DREAMPlace: Deep Learning Toolkit-Enabled GPU Acceleration for Modern VLSI Placement. In *Proceedings of the 56th Annual Design Automation Conference 2019 (DAC '19)*. Association for Computing Machinery, New York, NY, USA.
[15] Jingwei Lu, Pengwen Chen, Chin-Chih Chang, Lu Sha, Dennis Jen-Hsin Huang, Chin-Chi Teng, and Chung-Kuan Cheng. 2015. EPlace: Electrostatics-Based Placement Using Fast Fourier Transform and Nesterov's Method. 20, 2 (2015).
[16] Azalia Mirhoseini, Anna Goldie, Hieu Pham, Benoit Steiner, Quoc V Le, and Jeff Dean. 2018. A Hierarchical Model for Device Placement. In *ICLR*.
[17] Azalia Mirhoseini, Hieu Pham, Quoc V. Le, Benoit Steiner, Rasmus Larsen, Yuefeng Zhou, Naveen Kumar, Mohammad Norouzi, Samy Bengio, and Jeff Dean. 2017. Device Placement Optimization with Reinforcement Learning. In *ICML*.
[18] Volodymyr Mnih, Adrià Puigdomènech Badia, Mehdi Mirza, Alex Graves, Timothy P. Lillicrap, Tim Harley, David Silver, and Koray Kavukcuoglu. 2016. Asynchronous Methods for Deep Reinforcement Learning. arXiv:cs.LG/1602.01783
[19] OpenAI. [n.d.]. OpenAI Five. https://blog.openai.com/openai-five/.
[20] F. Scarselli, M. Gori, A. C. Tsoi, M. Hagenbuchner, and G. Monfardini. 2009. The Graph Neural Network Model. *IEEE Transactions on Neural Networks* 20, 1 (Jan 2009), 61–80. https://doi.org/10.1109/TNN.2008.2005605
[21] John Schulman, Sergey Levine, Philipp Moritz, Michael I. Jordan, and Pieter Abbeel. 2015. Trust Region Policy Optimization. arXiv:cs.LG/1502.05477
[22] John Schulman, Filip Wolski, Prafulla Dhariwal, Alec Radford, and Oleg Klimov. 2017. Proximal Policy Optimization Algorithms. arXiv:cs.LG/1707.06347
[23] Huang-A. Maddison C Silver, D. 2016. Mastering the game of Go with deep neural networks and tree search. *Nature* (2016).
[24] Oriol Vinyals, Igor Babuschkin, Junyoung Chung, Michael Mathieu, Max Jaderberg, Wojtek Czarnecki, Andrew Dudzik, Aja Huang, Petko Georgiev, Richard Powell, Timo Ewalds, Dan Horgan, Manuel Kroiss, Ivo Danihelka, John Agapiou, Junhyuk Oh, Valentin Dalibard, David Choi, Laurent Sifre, Yury Sulsky, Sasha Vezhnevets, James Molloy, Trevor Cai, David Budden, Tom Paine, Caglar Gulcehre, Ziyu Wang, Tobias Pfaff, Toby Pohlen, Dani Yogatama, Julia Cohen, Katrina McKinney, Oliver Smith, Tom Schaul, Timothy Lillicrap, Chris Apps, Koray Kavukcuoglu, Demis Hassabis, and David Silver. 2019. AlphaStar: Mastering the Real-Time Strategy Game StarCraft II. https://deepmind.com/blog/alphastar-mastering-real-time-strategy-game-starcraft-ii/.
[25] R.J. Williams. 1992. Simple statistical gradient-following algorithms for connectionist reinforcement learning. *Mach Learn* (1992).
[26] Zonghan Wu, Shirui Pan, Fengwen Chen, Guodong Long, Chengqi Zhang, and Philip S. Yu. 2019. A Comprehensive Survey on Graph Neural Networks. arXiv:cs.LG/1901.00596
[27] Jinjun Xiong, Yiu-Chung Wong, Egino Sarto, and Lei He. 2006. Constraint Driven I/O Planning and Placement for Chip-package Co-design. In *APSDAC*.
[28] Yanqi Zhou, Sudip Roy, Amirali Abdolrashidi, Daniel Wong, Peter C. Ma, Qiumin Xu, Ming Zhong, Hanxiao Liu, Anna Goldie, Azalia Mirhoseini, and James Laudon. 2019. GDP: Generalized Device Placement for Dataflow Graphs. arXiv:cs.LG/1910.01578